# A Benchmark and Comparison of Active Learning for Logistic Regression

**Yazhou Yang**  Y.YANG-4@TUDELFT.NL
**Marco Loog**  M.LOOG@TUDELFT.NL
**Pattern Recognition Laboratory, Delft University of Technology, The Netherlands**

## Abstract

Logistic regression is by far the most widely used classifier in real-world applications. In this paper, we benchmark the state-of-the-art active learning methods for logistic regression and discuss and illustrate their underlying characteristics. Experiments are carried out on three synthetic datasets and 44 real-world datasets, providing insight into the behaviors of these active learning methods with respect to the area of the learning curve (which plots classification accuracy as a function of the number of queried examples) and their computational costs. Surprisingly, one of the earliest and simplest suggested active learning methods, i.e., uncertainty sampling, performs exceptionally well overall. Another remarkable finding is that random sampling, which is the rudimentary baseline to improve upon, is not overwhelmed by individual active learning techniques in many cases.

## 1. Introduction

In practice, it is easy to acquire a large amount of data, yet difficult, time-consuming, and expensive to label data since human experts are usually involved (Settles, 2010). For instance, collecting millions of images from Google is not that difficult, while categorizing these images may need a lot of manpower and other resources. Active learning addresses this challenge by selecting the most valuable subset from the whole data set for human annotation. Many research studies have demonstrated that active learning is effective in maintaining good performance while reducing the overall labeling effort over a diverse range of applications, such as text categorization (Tong & Koller, 2002; Cai & He, 2012), medical image classification (Hoi et al., 2006; Saito et al., 2015), remote sensing (Tuia et al., 2012; Samat et al., 2016; Deng et al., 2017), image retrieval (Cheng & Wang, 2007; Liu et al., 2008; Zhang et al., 2010) and natural language processing (Tang et al., 2002).

To choose the most informative subset, it is of vital importance to choose an appropriate criterion which measures the usefulness of unlabeled instances. Most commonly used criteria in active learning include query-by-committee (Seung et al., 1992), uncertainty sampling (Tong & Koller, 2002), expected error minimization (Roy & Mccallum, 2001; Guo & Greiner, 2007; Holub et al., 2008), and variance reduction (Zhang & Oles, 2000; Yu et al., 2006; Schein & Ungar, 2007), variance maximization (Yang & Loog, 2018), maximum model change (Settles et al., 2008; Freytag et al., 2014; Käding et al., 2016; Cai et al., 2017) and the "min-max" view active learning (Hoi et al., 2008; Huang et al., 2014). They are derived from diverse heuristics and classifier dependent. Some of them are specifically designed for one particular classifier, *e.g.* the simple margin criterion for support vector machines (Tong & Koller, 2002), while others can be adapted to different types of classifiers, *e.g.* expected error reduction for logistic regression and naive Bayes (Roy & Mccallum, 2001).

In this work, we benchmark the state-of-the-art active learning algorithms built on logistic regression. Logistic regression is chosen because it is the most widely applied classifier in general and especially outside of machine learning in the applied sciences[1]. In addition, it is also used by most active learners (see, for instance, (Cuong et al., 2014; Gu et al., 2014; Guo & Greiner, 2007; Guo & Schuurmans, 2007; Hoi et al., 2006; 2009; Kanamori, 2007; Kanamori & Shimodaira, 2003; Liu et al., 2015; Schein & Ungar, 2007)). In part, the latter is because logistic regression readily provides an estimate of the posterior class probability, which is often exploited in active learning. In the binary classification setting, logistic regression models a posterior probability $P(y_i|x_i) = 1/(1 + \exp^{-y_i w^T x_i})$, where $x_i \in \mathbb{R}^d$ is a training feature vector labeled with $y_i \in \{+1, -1\}$ and $w$ is the $d$-dimensional parameter vector that is determined at training time. During training, we minimize the log-likelihood of the training data $\mathcal{L}$ to learn

---

[1] An advanced search on www.nature.com on October 1, 2017, gives us, for example, 1,126 hits for "support vector machine", 6,182 for "nearest neighbor" (containing more hits than just to the classifier), 1,231 for "LDA", and 14,715 for "logistic regression". Other classifiers are retrieved even less often.

the model parameter $w$ as follows:

$$\min_w \frac{\lambda}{2}\|w\|^2 + \sum_{x_i \in \mathcal{L}} \log(1 + \exp^{-y_i w^T x_i}) \quad (1)$$

where $\|w\|^2$ is a regularization term for which $\lambda$ controls its influence.

All in all, we study six different categories of active learning algorithms in which nine active learners are compared in an extensive benchmark study. Our work differs from two relevant earlier surveys on active learning (Settles, 2010; Fu et al., 2013) in two important respects: (1) our work constructs extensive experiments to investigate the empirical behaviors of these active learning algorithms while these two surveys do not compare the performance of different methods; (2) our paper presents a detailed summary of the active learning algorithms on the basis of logistic regression classifier because of its popularity while these two surveys offer an overview of existing active learning algorithms without specifying a type of classifiers. We believe that an empirical comparison can lead to a better understanding of the characteristics of active learning algorithms and provide guidance to the practitioner to choose a proper active learning algorithm. We should also mention the work by Schein & Ungar (2007) here, that already provided an evaluation of active learning methods using logistic regression. In this paper, however, we compare some new methods, which appeared only recently (Li & Guo, 2013; Huang et al., 2014; Cai et al., 2017), and we generally provide a fair and comprehensive comparison with much more extensively conducted experiments. We also investigate how active learning algorithms generally perform in comparison to random sampling, and point out the underlying relationships among the compared methods. The computational cost of each method is also evaluated.

In this paper, we focus on the pool-based setting, where few labeled samples and a large pool of unlabeled samples are available (Settles, 2010). We consider the myopic active learning which assumes that a single unlabeled instance is queried at a time. Batch mode active learning, which selects a batch of examples simultaneously, is not considered in this work and we refer to (Hoi et al., 2006; Guo & Schuurmans, 2007; Chakraborty et al., 2010; 2013; 2014; Chang & Liao, 2017) for further background of typical approaches.

The main contributions of this work can be summarized as follows:

1. A review of the state-of-the-art active learning algorithms built on logistic regression is presented, in which links and relationships between methods are explicated;
2. A preference map is proposed to reveal characteristic similarities and differences of the selection locations in 2D problems;
3. Extensive experiments on 44 real-world datasets and three artificial sets are carried out;
4. Insight is provided for the behaviors of classification performance and computational cost.

### 1.1. Outline

The remainder of the paper is organized as follows. Section 2 describes the general procedure of active learning and reviews the various approaches to active learning built on logistic regression. At the same time it sketches the relationships among different methods. Extensive experimental results on synthetic and real-world datasets are given in Section 3. The experimental setup is described and the outcomes are reported. More importantly, it provides an extensive discussion of the findings and aims to critically evaluate these compared methods. Section 4 concludes our work.

## 2. Active Learning Strategies and Methods

For myopic active learning in the pool-based scenario, we assume that a small set of labeled instances with a large pool of unlabeled samples are available. Let $\mathcal{L} = \{(x_i, y_i)\}_{i=1}^{l}$ represent the training data set that consists of $l$ labeled instances and let $\mathcal{U}$ be the pool of unlabeled instances $\{x_i\}_{i=l+1}^{n}$. Each $x_i \in \mathbb{R}^d$ is a $d$-dimensional feature vector and $y_i \in C$ is the class label of $x_i$. In this work we restrict ourselves to binary classification, which does not pose any essential limitation. For this reason, $C$ is simply taken to be the set $\{+1, -1\}$. The active learner will select an instance $x^*$ from the unlabeled pool based on its measure of utility, obtain the corresponding label $y^*$ by manual annotation and extend the training set with the new labeled sample $\mathcal{L} = \mathcal{L} \cup (x^*, y^*)$. The whole procedure is repeated until some stopping criteria are satisfied.

The remaining part of this section presents six different categories of active learning algorithms built on logistic regression, i.e., uncertainty sampling, error reduction, variance reduction, minimum loss increase, maximum model change and an adaptive approach, one per subsection. As also shown in Fig 1, nine different active learners which relate to the above six categories are used in our benchmark and comparison.

### 2.1. Uncertainty Sampling

Uncertainty sampling, which selects the instances for which the current classifier is least certain, is a widely used active learning method (Lewis & Gale, 1994; Settles, 2010). Querying these least certain instances can help the model refine the decision boundary. Intuitively, the distances between unlabeled instances and the deci-sion boundary can be measures of the uncertainty. Tong & Koller (2002) proposed a simple margin approach which queries the instance closest to the decision boundary.

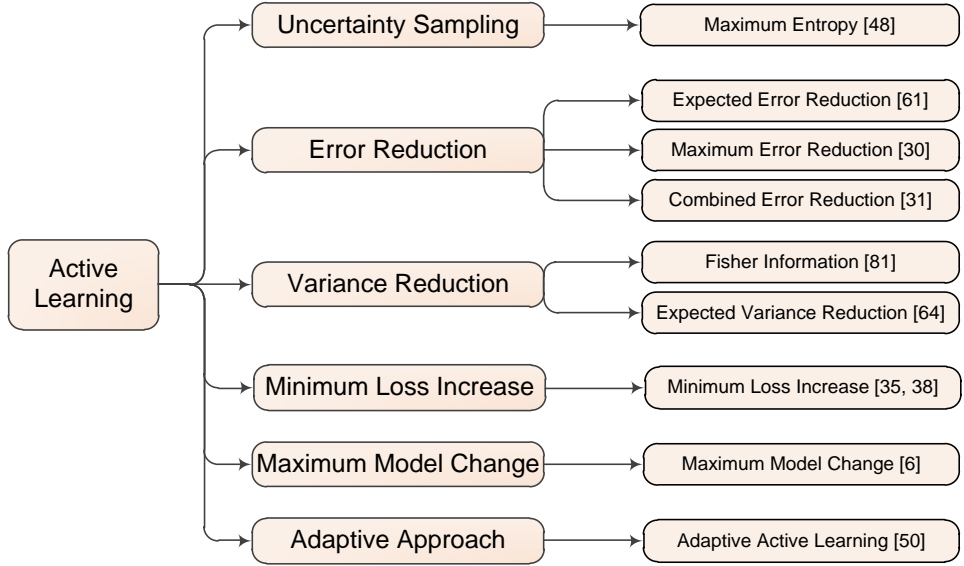

*Figure 1.* Nine active learners from six different categories are used in our comparison.

Entropy is a different and more widely used general measure of uncertainty (Shannon, 1948). Entropy-based approaches query the instances with *maximum entropy*:

$$x^* = \arg\max_{x \in \mathcal{U}} - \sum_{y \in C} P_{\mathcal{L}}(y|x) \log P_{\mathcal{L}}(y|x) \quad (2)$$

where $P_{\mathcal{L}}(y|x)$ is the conditional probability of $y$ given $x$ according to a logistic classifier trained on $\mathcal{L}$. This method is called ENTROPY for short. It calculates the entropy of each $x \in \mathcal{U}$ and selects the instance $x^*$ which has maximum entropy. It can be used with any classifier that produces probabilistic outputs. For binary classification, ENTROPY is equivalent to the simple margin approach (Tong & Koller, 2002).

One of the main risks of such uncertainty sampling based approaches lies in the fact that, due to a lack of exploration, they can get stuck at suboptimal solutions, continuously selecting instances which do not improve the current classifier at all (Huang et al., 2014).

### 2.2. Error Reduction

Error reduction approaches are another type of popular active learning methods (Roy & Mccallum, 2001; Guo & Greiner, 2007; Holub et al., 2008; Guo & Schuurmans, 2007). These approaches attempt to measure how much the generalization error is likely to be reduced when adding one new instance into the labeled dataset. Though one does not have direct access to the future test data, Roy & Mccallum (2001) proposed to estimate the expected error rate over the unlabeled examples under the assumption that the unlabeled data is representative of the test data. In other words, the unlabeled pool can be viewed as a validation set. Roy and Mccallum proposed to estimate the expected error using expected log-loss or 0-1 loss. For the former, which we consider in our work, the following objective is considered:

$$x^* = \arg\min_{x \in \mathcal{U}} \sum_{y \in C} P_{\mathcal{L}}(y|x) \left( -\sum_{x_i \in \mathcal{U}} \sum_{y_i \in C} P_{\mathcal{L}^+}(y_i|x_i) \log P_{\mathcal{L}^+}(y_i|x_i) \right) \quad (3)$$

Here, $\mathcal{L}^+ = \mathcal{L} \cup (x, y)$ indicates that the selected instance $x$ is labeled $y$ and added to the labeled dataset $\mathcal{L}$. We refer to this method as Expected Error Reduction (EER) in this paper. The first term $P_{\mathcal{L}}(y|x)$ is the posterior probability of $y$ given $x$ trained on the labeled dataset $\mathcal{L}$.

However, since using the labeled data $\mathcal{L}$, which is typically of small size, can result in a bad classifier, $P_{\mathcal{L}}(y|x)$ may not be estimated very adequately (Guo & Greiner, 2007). To avoid problems with such misspecifications, Guo & Greiner (2007) proposed an optimistic or, equivalently, *maximum* error reduction approach (called MAXER in this paper), which estimates the best-case error reduction, with-out considering $P_\mathcal{L}(y|x)$. MAXER considers the following objective instead:

$$x^* = \arg\min_{x \in \mathcal{U}} \min_{y \in C} \left( -\sum_{x_i \in \mathcal{U}} \sum_{y_i \in C} P_{\mathcal{L}^+}(y_i|x_i) \log P_{\mathcal{L}^+}(y_i|x_i) \right) \quad (4)$$

Note that the error reduction approaches above only take the unlabeled data into consideration when estimating the future error. To obtain better generalization performance, it has been suggested to compute the loss both over the training set $\mathcal{L}$ and over the unlabeled set $\mathcal{U}$. This idea was proposed in (Grandvalet & Bengio, 2004) for semi-supervised learning, while Guo & Schuurmans (2007) extended it to the batch mode active learning. Focusing on the myopic setting, one can adopt the related criterion as follows:

$$x^* = \arg\min_{x \in \mathcal{U}} \min_{y \in C} - \left( \sum_{x_j \in \mathcal{L}^+} \log P_{\mathcal{L}^+}(y_j|x_j) \right.$$
$$\left. + \alpha \sum_{x_i \in \mathcal{U} \setminus x} \sum_{y_i \in C} P_{\mathcal{L}^+}(y_i|x_i) \log P_{\mathcal{L}^+}(y_i|x_i) \right) \quad (5)$$

where $\alpha$ is a trade-off parameter used to adjust the importance of loss over labeled and unlabeled data. We name this combined approach CEER in this paper.

One general, potential disadvantage of error reduction approaches is the high computational cost (Settles, 2010). Each time a new queried instance $x$ with its label $y$ is added to the training dataset, we need to retrain the classifier to get the new posterior probabilities $P_{\mathcal{L}^+}(y_i|x_i)$. This retraining step may amount to great computational efforts.

### 2.3. Variance Reduction

Optimal experimental design, which attempts to minimize particular statistical criteria with the aim of saving in sampling cost, is an approach that has been classically used in the design of linear regression experiments (Atkinson et al., 2007; Yu et al., 2006; Lu et al., 2011). A-optimality, which is one of the classic, commonly used measures, is the trace of the inverse of the information matrix (Atkinson et al., 2007). Minimizing A-optimality can also be seen as reducing the average variance of the estimates of model parameters and therefore is wildly practised in active learning (Zhang & Oles, 2000; Hoi et al., 2006; Schein & Ungar, 2007).

In the binary classification setting, regarding regularized logistic regression, the Fisher information matrix over the unlabeled pool $\mathcal{U}$ is defined as $\mathcal{I}_\mathcal{U}(w) = \frac{1}{|\mathcal{U}|} \sum_{i \in \mathcal{U}} \sigma_i(1 - \sigma_i)x_i x_i^T + \lambda I_d$ where $\sigma_i = \sigma(w^T x_i) = 1/(1 + \exp(-w^T x_i))$ is the posterior probability of $P(y = 1|x_i)$, and $I_d$ is the identity matrix of size $d \times d$. Zhang & Oles (2000) utilized A-optimal design to minimize the Fisher information ratio between $\mathcal{I}_\mathcal{U}(\hat{w})$ and $\mathcal{I}_x(\hat{w})$:

$$x^* = \arg\min_{x \in \mathcal{U}} \text{tr}(\mathcal{I}_x(\hat{w})^{-1} \mathcal{I}_\mathcal{U}(\hat{w})) \quad (6)$$

where $\mathcal{I}_x(\hat{w}) = \sigma_i(1 - \sigma_i)x_i x_i^T + \lambda I_d$ and $\hat{w}$ is the maximum likelihood estimator. The entity $\mathcal{I}_\mathcal{U}(\hat{w})$ can be interpreted as the variance of model output with respect to unlabeled data $\mathcal{U}$, and $\mathcal{I}_x(\hat{w})^{-1} \mathcal{I}_\mathcal{U}(\hat{w})$ can be viewed as the future output variance once $x$ has been labeled. The criterion suggested selects unlabeled examples that minimize the Fisher information ratio or, equivalently, reduce the future variance. We call this approach Fisher information variance reduction (FIVR) in this paper. Hoi et al. (2006) exploited the same idea as in (Zhang & Oles, 2000) and extended it to the batch mode setting. When the batch size is set to one, Hoi's method is identical to FIVR apart from some approximations introduced for dealing with the batch setting.

Schein & Ungar (2007) proposed a similar A-optimal active learning method based on logistic regression. In doing so, one can define the Fisher information matrix over the training data $\mathcal{L}$ as $F = \frac{1}{l} \sum_{i \in \mathcal{L}} \sigma_i(1 - \sigma_i)x_i x_i^T + \lambda I_d$. Schein proposed to minimize the variance of the estimated distribution of the estimator $\sigma(\hat{w}^T x_i)$ as follows:

$$\text{Var}(\sigma(\hat{w}^T x_i)) \simeq c_i^T F^{-1} c_i$$

where $c_i = \sigma_i(1 - \sigma_i)x_i$ is the gradient vector of $\sigma_i$. The variance over all the unlabeled instances can be formulated as follows:

$$g(\mathcal{L}, \mathcal{U}) = \sum_{x_i \in \mathcal{U}} \sum_{y \in \{+1,-1\}} \text{Var}(\sigma(y\hat{w}^T x_i)) \simeq 2 \sum_{x_i \in \mathcal{U}} tr\{c_i^T F^{-1} c_i\}$$

The benefit of a newly selected instance, can then be measured in terms of the expected variance reduction:

$$x^* = \arg\min_{x \in \mathcal{U}} \sum_{y \in C} P_\mathcal{L}(y|x) g(\mathcal{L} \cup (x, y), \mathcal{U}) \quad (7)$$

We refer to this method as Expected Variance Reduction (EVR) in this paper. EVR represents the potential variance changes weighted by current estimated model $P_\mathcal{L}(y|x)$. EVR can also be extended to log-loss based EVR (Schein & Ungar, 2007), but we do not consider this algorithm any further since we observed that it generally performs poorer than EVR in our experiments.

EVR is similar to EER in some respects. First, see Eqs. (3) and (7), we can find that both EER and EVR measure the utility of an unlabeled instance $x$ by repeatedly labeling it $y$ (i.e. $y \in \{+1, -1\}$) and retraining the model on $\mathcal{L} \cup (x, y)$. Second, both of them calculate the expectation value, e.g. EER evaluates the expected future error while EVR computes the expected future variance.

EVR is also computationally expensive since it goes over all the pool and re-estimates $\hat{w}$ and Fisher information matrix $F$ each time. The need to calculate the inverse of matrix typically makes it even slower than expected error reduction approaches.

### 2.4. Minimum Loss Increase

The next heuristic we consider is minimum loss increase (MLI), which directly bases its criterion on already labeled samples. Related to this class, Hoi et al. (2008) originally proposed a min-max view of active learning that minimizes the gain of the objective function. We here look at the work of Hoi et al. (2008) in a more general formulation and demonstrate its relationship with the expected error reduction framework.

Let us consider an unconstrained optimization problem using an L2-loss regularized linear classifier and a loss function $V(w; x_i, y_i)$:

$$\min_{w} \quad g(w) = \frac{\lambda}{2}||w||^2 + \sum_{i=1}^{l} V(w; x_i, y_i) \quad (8)$$

where $y_i \in \{+1, -1\}$. Many loss functions can be adopted for linear classification. For example, hinge loss, $V(w; x_i, y_i) = \max(0, 1 - y_i w^T x_i)$, results in linear SVM and squared loss, $V(w; x_i, y_i) = (y_i - w^T x_i)^2$, leads to ridge regression. We will consider the logistic loss in the experimental section: $V(w; x_i, y_i) = \log(1 + \exp^{-y_i w^T x_i})$, which results in L2-regularized logistic regression.

Now, to identify the most valuable instances for labeling, we could select the example that, once labeled, results in the minimum gain in terms of the score of objective function. That is, we consider

$$x^* = \arg\min_{x \in \mathcal{U}} \max_{y \in C} g_{\mathcal{L}^+}(w) - g_{\mathcal{L}}(w) \quad (9)$$

where $\mathcal{L}^+ = \mathcal{L} \cup (x, y)$ and $g_{\mathcal{L}}(w)$ denotes the value of objective function over the training data $\mathcal{L}$. Since $g_{\mathcal{L}}(w)$ is independent of the next queried instance, we can rewrite Equation 9 as follows:

$$x^* = \arg\min_{x \in \mathcal{U}} \max_{y \in C} \min_{w} \frac{\lambda}{2}||w||^2 + \sum_{x_i \in \mathcal{L}^+} V(w; x_i, y_i) \quad (10)$$

This method can be interpreted as directly minimizing the worst-case value of the objective function when labeling a new instance. Considering kernel versions instead of linear classifiers in the above, they would entail the earlier mentioned min-max active learning methods (Hoi et al., 2008; Huang et al., 2014), which use the hinge loss and square loss, respectively. Hoi et al. (2008) originally presented the min-max view method and extended it to the batch mode active learning. Huang et al. (2014) extended the min-max view to consider all the unlabeled data and proposed an active learning method which QUeries Informative and Representative Examples (QUIRE for short) as follows:

$$x^* = \arg\min_{x \in \mathcal{U}} \min_{y_u \in C^{\{n_u-1\}}} \max_{y \in C} \min_{w} \frac{\lambda}{2}||w||^2 + \sum_{x_i \in \mathcal{L} \cup \mathcal{U}} V(w; x_i, y_i) \quad (11)$$

where $y_u$ indicates the labels of remaining unlabeled pool $\mathcal{U} \setminus x$ and $n_u$ is the number of samples of $U$. We need to point out, however, that the unlabeled part $\mathcal{U} \setminus x$ is of no use since QUIRE relaxed the constraints of $y_u$. Optimizing this unconstrained $y_u$ can guarantee that the remaining unlabeled data $\mathcal{U} \setminus x$ is useless, which can also be observed from Eqs. (9) and (10) in the original work (Huang et al., 2014). Therefore, QUIRE also fits this general framework.

As we consider the logistic loss for the above framework, MLI will refer to this particular choice. The particular objective function we consider is as follows:

$$\begin{aligned}
x^* &= \arg\min_{x \in \mathcal{U}} \max_{y \in C} \min_{w} \frac{\lambda}{2}||w||^2 + \sum_{x_i \in \mathcal{L}^+} V(w; x_i, y_i) \\
&= \arg\min_{x \in \mathcal{U}} \max_{y \in C} \min_{w} \frac{\lambda}{2}||w||^2 + \sum_{x_i \in \mathcal{L}^+} \log(1 + \exp^{-y_i w^T x_i}) \\
&= \arg\min_{x \in \mathcal{U}} \max_{y \in C} \frac{\lambda}{2}||\hat{w}||^2 + \sum_{x_i \in \mathcal{L}^+} -\log P_{\mathcal{L}^+}(y_i | x_i)
\end{aligned} \quad (12)$$

where $\hat{w}$ is the estimated parameter of the L2-regularized logistic regression model trained on the labeled data $\mathcal{L}^+ = \mathcal{L} \cup (x, y)$ and $P_{\mathcal{L}^+}(y_i|x_i) = 1/(1 + \exp^{-y_i \hat{w}^T x_i})$. Comparing Eqs. (5) and (12), we find that MLI differs from CEER in two respects: (1) MLI adopts the min-max criterion while CEER considers the best optimistic scenario (i.e. the smallest loss); (2) MLI only measures the log-likelihood on labeled data while CEER also takes the unlabeled data into account.

### 2.5. Maximum Model Change

Maximum mode change (MMC) is another strategy for active learning (Settles et al., 2008; Freytag et al., 2013; 2014; Käding et al., 2016; Cai et al., 2017). These approaches query the sample which can lead to a great change of the current model once labeled. The differences among these approaches lies in the criteria to measure the model change. Settles et al. (2008) proposed to measure the expected gradient length of the objective function. Freytag et al. (2014) estimated the change of model outputs instead of model pa-rameters.

Cai et al. (2017) proposed to use the gradient of the loss function to approximate the model change and derived algorithms for both SVM and logistic regression classifier. We briefly review this method built on logistic regression (Cai et al., 2017). Assumed that the loss of logistic regression after adding a new sample $(x, y)$ is

$$L(w) = - \sum_{i \in \mathcal{L}^+} \log(1 + \exp^{-y_i w^T x_i})$$

where $\mathcal{L}^+ = \mathcal{L} \cup (x, y)$ and $w$ is the parameter of logistic regression model. MMC approximates the model change as follows:

$$\frac{\partial L(w)}{\partial w} \approx \frac{\partial \log(1 + \exp^{-yw^T x})}{\partial w} = \frac{yx}{1 + \exp^{-yw^T x}}$$

Since the label $y$ is unknown, MMC calculates the expected model change

$$\mathcal{M}(x) = \mathbb{E}_y \left\| \frac{yx}{1 + \exp^{-yw^T x}} \right\| = \frac{2 \|x\|}{(1 + \exp^{-w^T x})(1 + \exp^{w^T x})} \quad (13)$$

Finally, MMC selects the sample $x^*$ that leads to the largest mode change as follows:

$$x^* = \arg\max_{x \in \mathcal{U}} \mathcal{M}(x) \quad (14)$$

Note that $\frac{1}{(1+\exp^{-w^T x})(1+\exp^{w^T x})}$ corresponds to $P(+1|x) \times P(-1|x)$. This value will be maximum when $P(+1|x) = 0.5$, which means that MMC prefers the sample with high uncertainty. In addition, MMC is also likely to query the instance with large norm. Therefore, MMC trades off the uncertainty and the norm of a sample.

### 2.6. Adaptive Active Learning

Li & Guo (2013) proposed an active learning approach which combines uncertainty sampling and information density measure in an adaptive way. We call this method Adaptive Active Learning (AAL). We should consider the instances which are located in a dense region for two reasons. One is that they are less likely to be the outliers. And secondly, they can represent the underlying distribution. By combining the uncertainty and information density measure, their proposed method can balance the informativeness and representativeness. There are some active learning methods that share a similar idea (Brinker, 2003; Settles & Craven, 2008; Zhu et al., 2010; Yang et al., 2015).

First, AAL trains a logistic regression classifier and uses the entropy as a measure of uncertainty, which is equivalent to the ENTROPY approach in Subsection 2.1. Then, AAL measures the information density by employing a Gaussian Process framework to calculate the mutual information between the candidate instance and the unlabeled pool. Finally, it combines the two criteria using a trade-off parameter $\beta$ ($0 \leq \beta \leq 1$):

$$h_\beta(x_i) = u(x_i)^\beta \times d(x_i)^{1-\beta} \quad (15)$$

where $u(x_i)$ and $d(x_i)$ are the uncertainty and density values of $x_i \in \mathcal{U}$, respectively.

It is difficult, however, to set a proper weighting parameter $\beta$. Instead of using a pre-defined value of $\beta$, Li & Guo (2013) proposed to adaptively choose the $\beta$ value from a given set $[0.1, 0.2, \ldots, 0.9, 1]$. Each different $\beta$ leads to picking a candidate instance from unlabeled samples. Among these candidates, AAL chooses the sample which has minimal expected classification error according to expected error reduction method (Roy & Mccallum, 2001). In other words, AAL adaptively changes the $\beta$ value to form a candidate set, from which the most informative sample is selected by using EER.

## 3. Experiments

The experimental setup is first described, followed by an analysis of the results on synthetic datasets and real-world datasets, respectively. Finally, we investigate the computational costs of different active learning algorithms.

### 3.1. Experimental Setting

We present the necessary information of three synthetic datasets and 44 real-world datasets that we used in the following subsections, along with a description of the evaluation design.

#### 3.1.1. SYNTHETIC DATA SETS

Three binary synthetic datasets are constructed to intuitively demonstrate the different behaviors of active learning algorithms. The first dataset *Synth1* is a standard *2D binary* problem which is shown in Fig 2a. Positive and negative classes are generated according to two multivariate normal distributions centered at $[1, 1]^T$ and $[-1, -1]^T$, respectively. We want to explore which active learning method works well on this unambiguously specified problem. The second dataset, *Synth2*, displayed in Fig 2c, is generated according to the description in (Huang et al., 2014). We can observe that Synth2 has a clear cluster structure. On this kind of data, uncertainty sampling has substantial problems since it only considers the most *uncertain* instance which comes closest to the decision boundary. Initially, the decision boundary estimated from the limited number of labeled data may be far away from the actual boundary and therefore uncertainty sampling may select less informative instances due to a poorly estimated posterior probability. This is exactly what this dataset was designed for and set out to illustrate. This dataset prefers some kind of active learning methods which can consider the so-called representativeness along with the informativeness at the same time (Huang et al., 2014). Representative instances are those that drive exploration, and not exploitation. The latter is what uncertainty sampling typically aims for. The third synthetic dataset, named *Synth3*, is also a 2D classification problem which is shown in Fig 2e. Synth3 is constructed to have a shape which looks like a tilted ⊔. Each part is generated from two multivariate normal distributions with small overlap. Compared with Synth1, Synth3 is a more challenging dataset since it has relatively complex structure and may mislead some active learning methods. We are curious whether active learning can outperform random sampling on this kind of data. We investigate how active learning approaches work in the above three synthetic datasets and whether they perform better than random sampling.

3.1.2. REAL-WORLD DATA SETS

As real-world benchmarks, we use various UCI datasets (Lichman, 2013), the MNIST handwritten digit dataset (LeCun et al., 1998), the 20 Newsgroups dataset (Lang, 1995) and the 80 subsets of the ImageNet database (Deng et al., 2009). Table 1 lists the preprocessed datasets used in our study together with some basic information. All the datasets are pre-processed to become binary classification problems.

There are a total of 44 benchmark datasets used in this comparison, including the ImageNet dataset on which extensive experiments on 80 binary subsets are conducted. Most datasets are pre-processed to have zero mean and unity standard deviation according to (Fernández-Delgado et al., 2014). Some datasets are linearly scaled to $[-1, 1]$ or $[0, 1]$ according to (Chang & Lin, 2011) [2]. These datasets have various sample sizes and diverse feature dimensionalities. Some of them can be quite easily handled while others are quite difficult classification problems. The Letter Recognition Data Set from UCI, which consists of 20,000 examples of 26 uppercase letters in various fonts and distortions, is also used as a test bed in (Frey & Slate, 1991). As in this last work, 16 attributes are extracted from the letters as the feature and we consider the following six classification tasks between pairs of letters: D vs. P, E vs. F, I vs. J , M vs. N, V vs. Y, and U vs. V. These pairs of letters are selected since they have a somewhat similar appearance and distinguishing them is challenging.

The MNIST [3] contains 60,000 training examples and 10,000 test examples which have been pre-processed to the same size of $28 \times 28$ pixels. The pairs 3 vs. 5, 5 vs. 8, and 7 vs. 9 constitute three difficult classification tasks and are used as the binary sets in our benchmark. For each of the three pairs, we randomly subsample 1500 instances from the original dataset for computational reasons. Each pixel value is extracted as a feature, resulting in a 784-dimensional feature.

The 20 Newsgroups is a common benchmark used for document classification [4]. We use one version of this dataset which consists of 18,846 instances of 20 different news topics. Similar to the work of (Zhu et al., 2003), our work also evaluates three binary tasks from this dataset: *sport.baseball* vs. *sport.hockey*, *pc.hardware* vs. *mac.hardware*, and *talk.religion.misc* vs. *alt.atheism*. All the documents have been pre-processed into 26,241 dimensional tf.idf vectors to which we initially apply PCA to reduce the dimensionality to 500 for computational reasons.

In addition, we also compare these active learning algorithms on a total of 80 binary subsets taken from the large visual ImageNet database (Deng et al., 2009). First, following the work of (Cai et al., 2017), we take 8 different subsets of ImageNet: five categories of cats (i.e. Egyptian, Persian, Siamese, Tabby and Tiger) and elephant, rabbit and panda. Subsequently, we construct eight binary-class classification problems by considering cat vs. elephant, cat vs. rabbit, cat vs. panda and each category of cat vs. the four remaining cats. Moreover, we also randomly chose 72 paired classes to generate 72 binary data sets from the ImageNet database provided by Tommasi & Tuytelaars (2014). SIFT features are first extracted and then encoded into 1000-bin histograms. Detailed information of the 80 subsets of the ImageNet dataset is included in the Appendix.

3.1.3. EVALUATION DESIGN

In the evaluation, each dataset is randomly divided into training and test data sets of equal size. Following some previous work (McCallumzy & Nigamy, 1998; Zhu et al., 2003; Tong & Koller, 2002; Guo, 2010; Gu et al., 2014; Li et al., 2012), we consider a difficult case of active learning, where only two labeled instances are provided as the initial labeled set, one from each class. We repeat each experiment 20 times on each real-world dataset. As for the synthetic datasets, we repeat the experiments 1000 times and every time we randomly regenerate the whole dataset. The average performance of each active learning method on each dataset is reported. In all the experiments, regularized logistic regression included in LIBLINEAR package (Fan et al., 2008) is used as the classifier. We set the regularization parameter $\lambda$ to 0.01. The trade-off parameter

---

[2] https://www.csie.ntu.edu.tw/ cjlin/libsvmtools/datasets/
[3] http://yann.lecun.com/exdb/mnist/

[4] http://qwone.com/ jason/20Newsgroups/

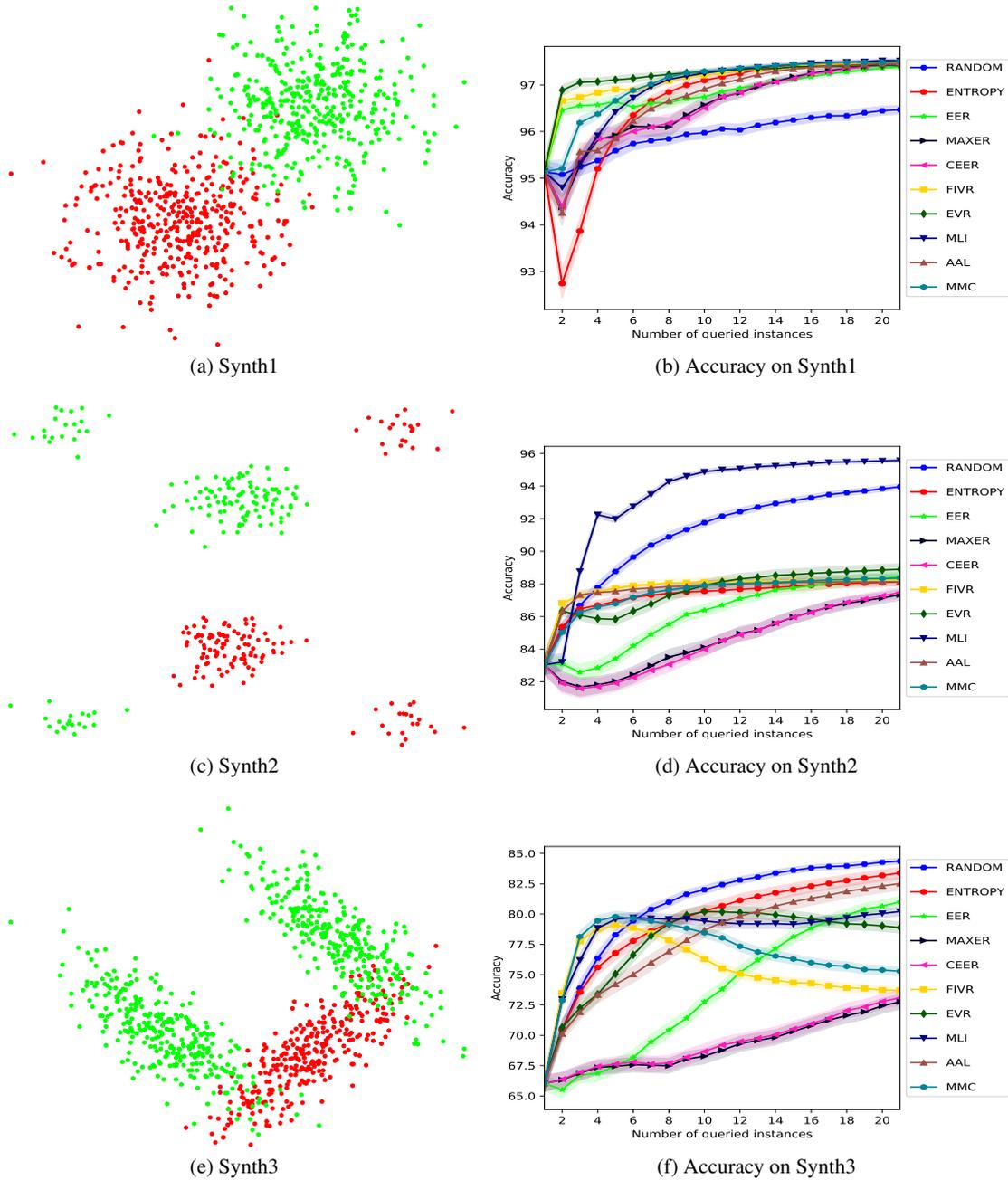

Figure 2. Distribution for three synthetic datasets and the results on these same sets in terms of classification accuracy with 90% confidence interval. Red and green points represent the two different classes. (a) shows the distribution of the Synth1 dataset; (b) presents the average accuracy of each active learning method on the test set for Synth1. (c)-(d) and (e)-(f) represent the same results for Synth2 and Synth3, respectively.

Table 1. Data sets information: It shows the number of instances (# INS) and the feature dimensionality (# FEA)

| Data set | # Ins | # Fea | Data set | # Ins | # Fea |
|---|---|---|---|---|---|
| ac-inflam | 120 | 6 | acute | 120 | 6 |
| australian | 690 | 14 | blood | 748 | 4 |
| breast | 683 | 10 | credit | 690 | 15 |
| cylinder | 512 | 35 | diabetes | 768 | 8 |
| fertility | 100 | 9 | german | 1000 | 24 |
| haberman | 306 | 3 | heart | 270 | 13 |
| hepatitis | 155 | 19 | hill | 606 | 100 |
| ionosphere | 351 | 34 | liver | 345 | 6 |
| mushrooms | 1000 | 112 | mammographic | 961 | 5 |
| musk1 | 476 | 166 | ooctris2f | 912 | 25 |
| ozone | 1000 | 72 | parkinsons | 195 | 22 |
| pima | 768 | 8 | planning | 182 | 12 |
| sonar | 208 | 60 | splice | 1000 | 60 |
| tictactoe | 958 | 9 | vc2 | 310 | 6 |
| vehicle | 435 | 18 | wisc | 699 | 9 |
| wdbc | 569 | 31 | letterDP | 1608 | 16 |
| letterEF | 1543 | 16 | letterIJ | 1502 | 16 |
| letterMN | 1575 | 16 | letterVY | 1577 | 16 |
| letterUV | 1550 | 16 | 3 vs 5 | 1500 | 784 |
| 5 vs 8 | 1500 | 784 | 7 vs 9 | 1500 | 784 |
| baseball vs hockey | 1993 | 500 | pc vs mac | 1945 | 500 |
| misc vs atheism | 1427 | 500 | subsets of ImageNet | 180821 | 1000 |

present in the active learners considered, $\alpha$ for CEER, is set to 1.

For performance comparison, classification accuracy (or equivalently, the error rate) is the defacto standard evaluation criterion: the higher the accuracy, the better the algorithm. In active learning, however, performance varies depending on the number of labeled samples that one is allowed to take and we cannot settle on a single number of added labeled samples. For this reason, we use the area under the learning curve (ALC) (Cook & Krishnan, 2015) as the evaluation criterion. The larger this value, the better the performance. The optimal score is 1.

### 3.2. Analysis on synthetic datasets

In Fig 2, we display the distributions of the three synthetic datasets, along with the performance of each active learning method in terms of the classification accuracy on the test set. We also present the 90% confidence interval around each learning curve. To start with, note that no single method outperforms all the other methods on all the datasets.

#### 3.2.1. PREFERENCE MAP

To generally show a difference in characteristic of the various active learning methods, we introduce a visualization technique, called Preference Map, for our synthetic datasets (see Fig 3 and 4).

The preference map is generated by keeping track of the locations of the queried instances selected by each active learning algorithm. Presenting kernel density plots of all these locations and displaying them using pseudo-colors gives an impression where in feature space the active learners request their data from. The highest density regions are marked in red while the lowest density regions are indicated in blue. The preference map of the instance first queried is shown in Fig 3a. More specifically, for our 2D synthetic datasets, we record the location of the first queried sample selected by each active learner during 1000 repetitions of the experiment and generate the density plots.

We also plot the preference maps corresponding to the complete learning, where we exponentially weigh down later observations based on the intuition that the examples selected early on in the process are more valuable than the examples selected in the later rounds. The specific weight function we employ is $\exp(-r/R)$, where $r$ and $R$ are the current round and the total rounds, respectively. In other words, we make a record of the locations of all queried samples during the whole active learning process, followed by producing weighted preference maps. The corresponding preference maps are in Fig 3b.

### 3.2.2. RESULT ON SYNTH1

Synth1 is a simple classification problem and some algorithms perform well in the beginning stage, such as the variance reduction approaches FIVR and EVR. On the other hand, ENTROPY achieves rather poor performance at the beginning and is the worst approach at the first selected point in Fig 2b. To understand this specific aspect of how uncertainty sampling behaves, we refer to the preference map in Fig 3a. We can see that random sampling prefers the region where the mean vector of each class is. Clearly, the preference map for random sampling should ultimately reproduce the original underlying distribution, which is a mixture of two normal distributions in the setting we consider. Uncertainty sampling clearly prefers to query the points in the middle of two clusters since it focuses on the instances near the estimated decision boundary. Even though these samples may be close to the true decision boundary, they may not be a good choice, as they lead to instable estimates. This is what we see in the results, where ENTROPY performs rather poor in the beginning stage. CEER and MAXER show similar behaviors in the preference map and also seem to give relatively worse performance at the start of the active learning cycle. Their maps, however, seem a bit more rectangular, which may lead to slightly improved stability and therefore better performance as compared to ENTROPY. Variance reduction methods like EVR and FIVR also sample parallel to the decision boundary, but more through the respective class centers, which indeed leads to more stable and therefore better performing estimates. MLI, on the other hand, seems to sample perpendicular to the decision boundary, away from the regions with high density. This may be because MLI, which is similar to QUIER (Huang et al., 2014), tends to balance the informativeness and representativeness. When only two initial points are available, MLI prefers to select the instance far away from already labeled ones. AAL queries the first instance from a broad region since it is able to explore a large region by adaptively changing the tradeoff parameter. MMC performs similarly to MLI. The reason may be that MMC balances uncertainty and the norm of unlabeled instance. MMC prefers the samples with large norm and high uncertainty.

Turning our attention to the overall weighted preference maps Fig 3b, we see a dramatic change in behavior for at least six strategies. FIVR and especially EVR change their sampling from parallel to more perpendicular to the decision boundary. The changes we see for EER and MLI may be interpreted as changes from the more explorative initial phase to a more exploitative later stage, where a sampling around the decision boundary is performed to refine it. That active learning should actually deal with the exploration-exploitation tradeoff is at the basis of MLI. AAL also changes from the initial exploration to the subsequent exploitation. MMC seems to attach more importance to the uncertainty than the norm of sample. In addition, we observe that some active learning approaches, of which overall preference maps are similar to each other, performs similarly to each other in the later stage (e.g. after 8 samples are queried). For example, FIVR and MMC have similar maps while their performances are almost identical when 8 instances are labeled.

### 3.2.3. RESULT ON SYNTH2

Fig 2d shows that on the second synthetic problem, random sampling far surpasses all the active learning methods except for MLI, which is the best performing strategy. We use the overall preference map to explain this result. Fig 4 displays preference maps corresponding to the whole learning curve on Synth2 dataset.

We can see that the preference map of random sampling reproduces the underlying distribution. The preference maps of the remaining active learning methods except MLI, which are almost identical to each other, only highlight the two large clusters in the middle. This indicates that most of the queried samples are from the two middle clusters. This happens because that these active learning methods are misled by an incorrect model estimated with limited initial training samples. For instance, assume that we have two initial labeled points separately located in the two middle clusters. This initial training data will lead to a completely wrong estimation of the decision boundary. Then, these active learners will keep selecting the points which come close to the wrong decision boundary. However, these selected points cannot provide much more information about the true underlying distribution. Finally, they miss a chance of selecting the samples from other small clusters to discover the underlying distribution. This shows a common situation where some active learning methods get stuck in keeping querying useless instances due to inaccurate estimation of model parameter. Random sampling does not suffer from this because it acts purely random in selecting new instances. This is why random sampling surpasses these active learning methods.

MLI can perform even better in this situation. MLI can select the samples in the upper left corner and lower right corner on Synth2 dataset since it also considers the so-called representativeness of each instance, such as whether the instances are inside of some clusters (Huang et al., 2014). This leads to the exploratory behavior of MLI. As shown in Fig 4, MLI is more likely to query the instances along four clusters on the border line while some methods like uncertainty sampling and error reduction approaches favor the instances in the two middle clusters. This is the reason that MLI can significantly outperform random sampling and other active learning methods on this artificial set.

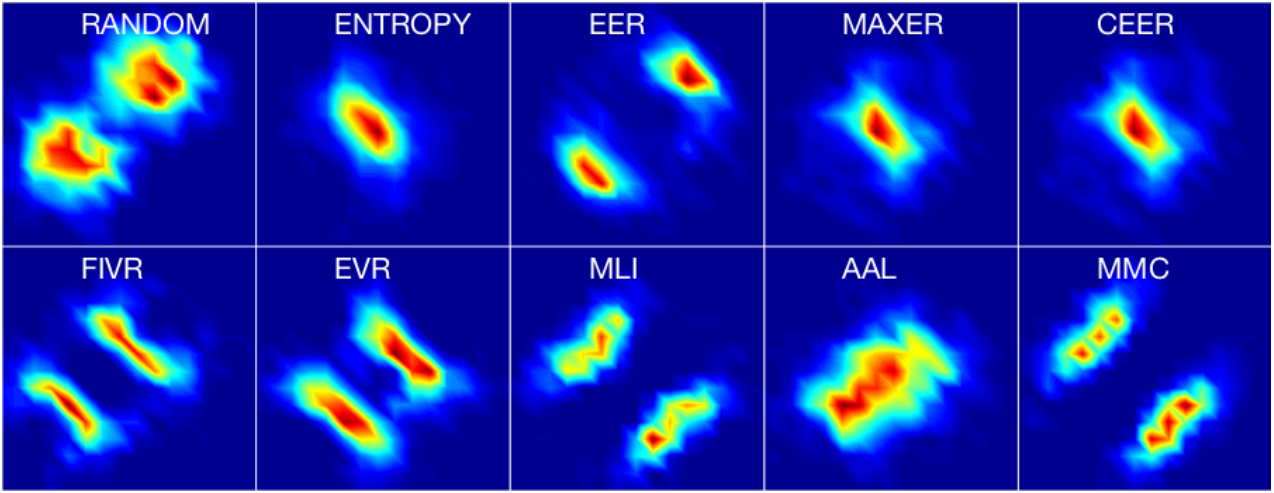

(a)

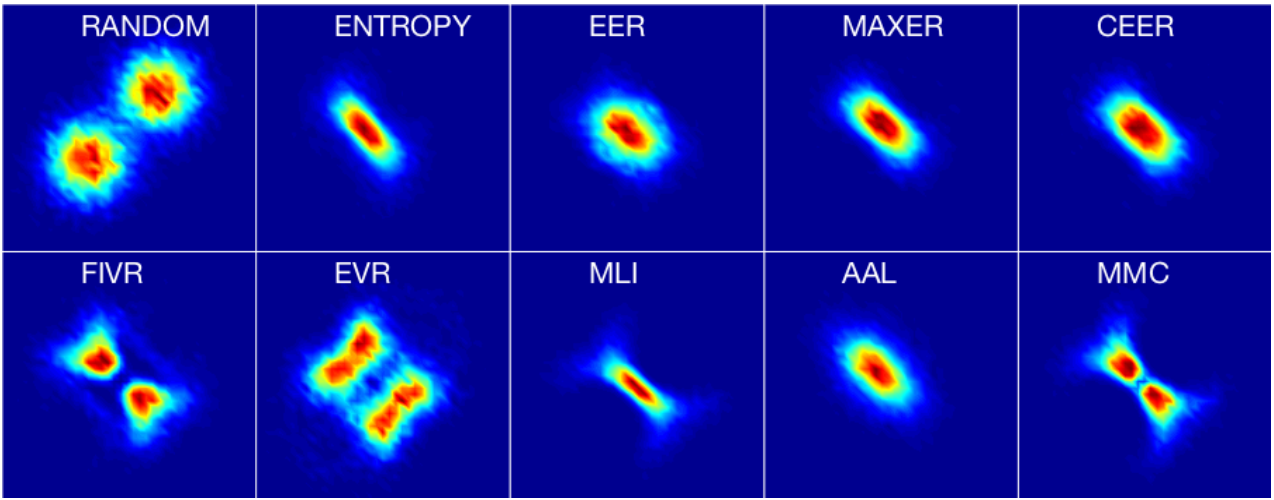

(b)

*Figure 3.* (a) Preference maps of first queried example selected by each active learning method on the Synth1 dataset; (b) Weighted preference maps over the whole learning process on the same problem.

### 3.2.4. RESULT ON SYNTH3

From Fig 2f, we can observe some negative results that random sampling outperforms all the other active learning methods after 6 instances are selected. The possible reason is that random sampling can explore the whole structure of this dataset while other methods just pay attention to some local parts without exploring the whole dataset. And another reason may be that it is difficult to achieve good classification result on this dataset due to its complex structure. On this kind of hard datasets, active learning methods can easily get stuck in local structure while ignoring the global view of the problem. Due to space limitations, the preference maps of Synth3 are omitted.

### 3.3. Analysis on real-world datasets

Table 2 presents the results for applying each active learning method on the real-world datasets. We adopt the paired $t$-test at a $95\%$ significance level on all the experiments to test which method does not significantly differ from the best method. The best performance is highlighted in bold face and surrounded with a box, together with the competitors that perform at a comparable level. The av-erage ALC ("Mean" in Table 2) of each method is also reported. "Average Ranking" shows the average ranking of compared methods. "Win counts" shows the total number of datasets on which one method achieves the best performances. "win/tie/loss" demonstrates the win, tie, and loss counts of one method versus random sampling over all of the datasets.

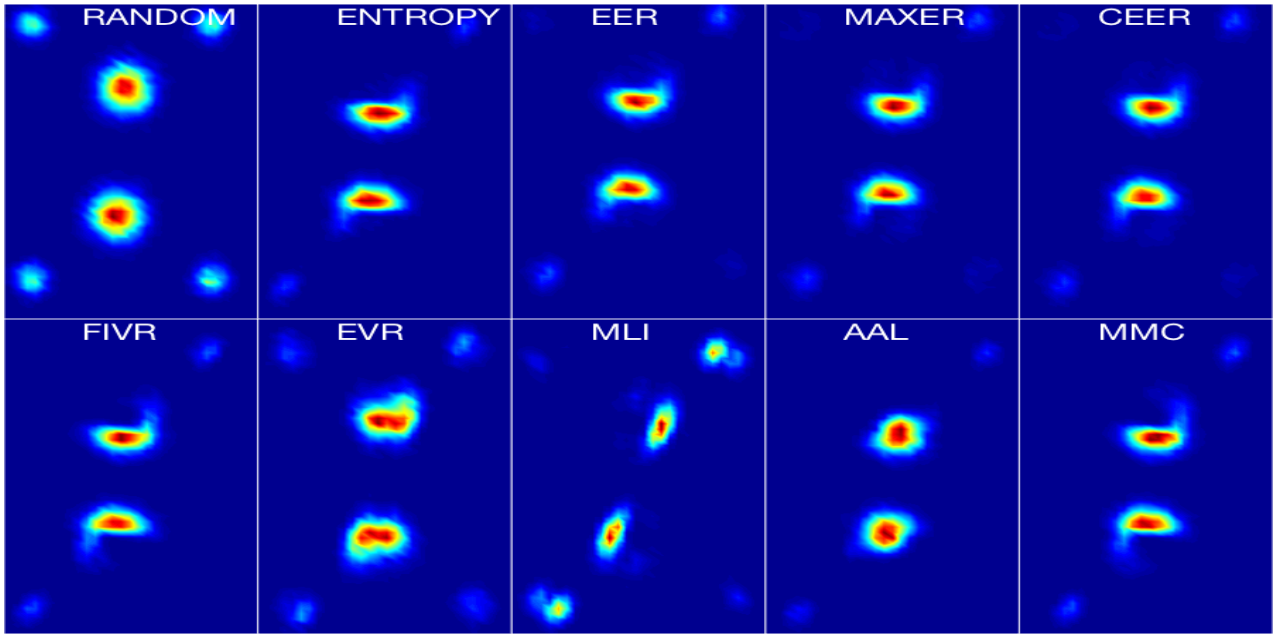

*Figure 4.* Weighted preference maps for the overall learning procedure by each of the active learning methods on the Synth2 dataset.

As shown in Table 2, no single algorithm outperforms all others on all the datasets. Still ENTROPY and EER seem to markedly outperform other active learning methods. It may be surprising that uncertainty sampling can compete with relatively sophisticated active learning algorithms as it is a rather simplistic approach. In fact, uncertainty sampling gets the highest ALC score and performs best in terms of win/tie/loss counts versus random sampling. It also obtains the best average ranking. MLI and MMC behave the second best among the remaining methods in terms of win counts while their average ALC and average ranking are outperformed by uncertainty sampling.

Considering the error reduction approaches, it is clear that EER outperforms MAXER and CEER. The overall performances of MAXER and CEER remain close to that of random sampling. MAXER merely surpasses random sampling on 20 of the 44 datasets. This seems to demonstrate that the best-case criterion is not an appropriate choice for active learning, at least for error reduction approaches. The possible cause may be that such optimistic measure simply puts too much trust in a typically badly estimated model. As a result, initial errors may get reinforced rather than mitigated by correctly chosen additional samples. This is comparable to some of the issues that arise in self-learning and EM-based approaches to semi-supervised learning (Loog & Jensen, 2015; Cohen et al., 2004). Guo & Greiner (2007) proposed the on-line adjustment for MAXER, which switches to another active learning method when MAXER supposedly guesses wrong about the true label of latest queried instance. We do not adopt this adjustment since it can be used for any active learning algorithms and we only focus on the performance of original, pure active learning methods. CEER obtains performance comparable to that of MAXER, and it shares the same problem with MAXER since it also uses the optimistic strategy (Guo & Schuurmans, 2007). One possible reason why CEER underperforms is that the trade-off parameter $\alpha$ is not well determined.

As for the variance reduction approaches, EVR slightly outperforms FIVR in terms of average scores and win/tie/loss counts. While FIVR achieves better performance than MAXER and CEER, it is still exceeded by random sampling on 12 datasets. EVR behaves comparably to EER. Three remaining methods, MLI, AAL and MMC, have similar performances on average ALC score. ENTROPY, MLI and MMC perform the best in terms of win counts. However, MMC is surpassed by random sampling on 14 datasets. AAL performs worse than random sampling on 17 datasets.

As random sampling is the technique to beat, it is important to see how the active learners perform in comparison

# A Benchmark and Comparison of Active Learning for Logistic Regression

*Table 2.* Performance comparison on the Area under the Learning Curve (Accuracy). The higher the score, the better the performance. For each data set, the best performances and its comparable competitors are highlighted in bold face and surrounded with a box. Average performance of all the active learning methods are also reported as "Mean". "Average Ranking" shows the average ranking of compared methods. "Win counts" shows the total number of datasets on which one method achieves the best performances. "win/tie/loss" demonstrates the win/tie/loss counts of one method versus random sampling on all the datasets based on paired $t$-test at 95 percent significance level.

| Dataset | Random | ENTROPY | EER | MAXER | CEER | FIVR | EVR | MLI | AAL | MMC |
|---|---|---|---|---|---|---|---|---|---|---|
| hill | 0.581 | 0.593 | 0.616 | 0.606 | 0.575 | 0.549 | 0.590 | **0.626** | 0.544 | 0.570 |
| planning | 0.586 | 0.584 | 0.580 | 0.568 | 0.574 | 0.574 | 0.587 | **0.614** | 0.575 | 0.581 |
| cylinder | 0.586 | 0.593 | 0.610 | 0.575 | 0.576 | **0.630** | 0.591 | 0.608 | 0.590 | 0.607 |
| liver | 0.627 | 0.612 | 0.635 | 0.623 | 0.621 | **0.645** | 0.632 | 0.615 | 0.616 | 0.631 |
| splice | 0.659 | **0.685** | 0.679 | 0.666 | 0.663 | 0.676 | 0.664 | 0.650 | 0.672 | 0.650 |
| german | 0.664 | 0.694 | 0.673 | 0.652 | 0.654 | 0.687 | 0.678 | 0.691 | 0.679 | **0.707** |
| ooctris2f | 0.679 | 0.665 | 0.678 | 0.648 | 0.652 | 0.651 | 0.680 | **0.686** | 0.669 | 0.646 |
| musk1 | 0.682 | **0.702** | **0.699** | 0.672 | 0.668 | 0.679 | 0.689 | **0.702** | 0.684 | 0.675 |
| fertility | 0.693 | 0.701 | 0.706 | 0.679 | 0.674 | 0.716 | 0.686 | **0.727** | 0.699 | 0.705 |
| haberman | **0.711** | **0.712** | **0.712** | 0.704 | 0.704 | 0.692 | 0.708 | 0.694 | **0.710** | 0.691 |
| sonar | 0.713 | 0.711 | 0.715 | 0.713 | 0.715 | 0.720 | 0.720 | 0.708 | **0.723** | 0.718 |
| pima | 0.716 | 0.717 | 0.706 | 0.710 | 0.707 | **0.727** | 0.708 | 0.711 | 0.709 | 0.700 |
| pcmac | 0.717 | 0.727 | 0.715 | 0.698 | 0.696 | 0.717 | 0.711 | 0.747 | 0.713 | **0.776** |
| diabetes | 0.719 | 0.721 | 0.723 | 0.726 | 0.724 | 0.707 | 0.725 | 0.726 | 0.709 | **0.736** |
| religionatheism | 0.720 | **0.740** | 0.710 | 0.687 | 0.687 | 0.712 | 0.704 | 0.691 | 0.702 | 0.720 |
| hepatitis | 0.731 | **0.753** | 0.753 | 0.744 | 0.741 | 0.745 | **0.754** | 0.730 | **0.757** | 0.708 |
| blood | **0.743** | 0.718 | 0.740 | 0.732 | 0.736 | 0.723 | 0.732 | 0.730 | 0.728 | 0.728 |
| heart | 0.774 | 0.793 | 0.791 | 0.788 | 0.788 | **0.799** | 0.781 | 0.797 | 0.784 | 0.787 |
| ImageNet | 0.778 | **0.783** | 0.763 | 0.761 | 0.760 | 0.775 | 0.765 | 0.762 | 0.761 | 0.774 |
| ionosphere | 0.779 | 0.782 | **0.818** | 0.806 | 0.801 | 0.790 | 0.812 | 0.674 | 0.812 | 0.768 |
| credit | 0.779 | **0.822** | 0.793 | 0.795 | 0.804 | **0.819** | 0.791 | 0.797 | 0.758 | 0.780 |
| mammographic | 0.780 | 0.779 | 0.774 | 0.781 | 0.784 | **0.795** | 0.777 | 0.766 | 0.775 | 0.775 |
| basehockey | 0.793 | 0.822 | 0.784 | 0.772 | 0.770 | 0.820 | 0.783 | 0.817 | 0.768 | **0.857** |
| vc2 | 0.807 | 0.814 | 0.815 | 0.802 | 0.803 | 0.816 | 0.822 | **0.825** | 0.796 | 0.823 |
| parkinsons | 0.811 | 0.823 | 0.824 | 0.824 | 0.828 | 0.825 | 0.825 | 0.830 | 0.803 | **0.838** |
| australian | 0.823 | **0.844** | 0.832 | 0.839 | 0.838 | 0.817 | 0.831 | 0.842 | 0.829 | 0.828 |
| letterIJ | 0.853 | 0.871 | 0.879 | 0.807 | 0.806 | 0.841 | 0.869 | 0.865 | 0.865 | **0.889** |
| letterVY | 0.855 | 0.880 | 0.878 | 0.814 | 0.814 | 0.753 | **0.886** | 0.861 | 0.876 | 0.830 |
| 3vs5 | 0.856 | 0.884 | **0.903** | 0.890 | 0.889 | 0.869 | **0.903** | 0.859 | 0.894 | 0.884 |
| vehicle | 0.859 | **0.884** | 0.878 | 0.851 | 0.855 | 0.830 | **0.884** | **0.883** | 0.837 | 0.847 |
| 5vs8 | 0.864 | 0.891 | **0.907** | 0.896 | 0.895 | 0.875 | **0.909** | 0.850 | 0.899 | 0.891 |
| 7vs9 | 0.876 | 0.904 | 0.914 | 0.905 | 0.906 | 0.909 | **0.917** | 0.841 | 0.912 | 0.904 |
| ozone | 0.882 | 0.884 | 0.860 | 0.862 | 0.861 | 0.843 | **0.901** | 0.892 | 0.863 | 0.868 |
| tictactoe | 0.894 | 0.902 | **0.912** | 0.673 | 0.684 | 0.903 | 0.898 | 0.853 | 0.902 | 0.843 |
| letterMN | 0.916 | **0.939** | **0.944** | 0.910 | 0.910 | 0.932 | 0.941 | 0.927 | 0.930 | 0.935 |
| mushrooms | 0.931 | **0.971** | 0.969 | 0.967 | 0.967 | **0.972** | 0.960 | **0.971** | 0.968 | **0.971** |
| letterEF | 0.933 | **0.959** | 0.954 | 0.949 | 0.950 | 0.954 | 0.956 | 0.956 | 0.952 | **0.957** |
| wdbc | 0.938 | 0.955 | 0.953 | 0.955 | 0.954 | 0.943 | 0.951 | **0.958** | 0.948 | 0.954 |
| letterDP | 0.939 | 0.964 | 0.963 | 0.950 | 0.950 | 0.954 | 0.961 | **0.967** | 0.956 | **0.967** |
| letterUV | 0.945 | 0.970 | **0.972** | 0.955 | 0.955 | 0.963 | 0.966 | **0.974** | 0.963 | **0.975** |
| wisc | 0.949 | 0.956 | 0.951 | **0.958** | 0.957 | 0.954 | 0.953 | 0.956 | 0.956 | 0.956 |
| breast | 0.950 | 0.958 | 0.956 | 0.956 | 0.957 | 0.960 | 0.955 | **0.962** | 0.947 | **0.962** |
| ac-inflam | 0.955 | **0.985** | 0.981 | 0.962 | 0.965 | 0.982 | 0.967 | 0.980 | **0.983** | **0.983** |
| acute | 0.977 | **0.991** | 0.971 | 0.958 | 0.965 | **0.991** | 0.954 | **0.992** | 0.986 | **0.991** |
| Mean | 0.796 | **0.810** | 0.809 | 0.791 | 0.790 | 0.801 | 0.806 | 0.803 | 0.800 | 0.804 |
| Average Ranking | 6.86 | **3.89** | 4.36 | 6.89 | 6.89 | 5.36 | 4.84 | 4.70 | 6.11 | 5.09 |
| Win counts | 2 | **14** | 8 | 1 | 0 | 8 | 7 | 13 | 4 | 13 |
| win/tie/loss | - | **33/7/4** | 32/3/9 | 20/5/19 | 21/3/20 | 29/3/12 | 32/3/9 | 30/2/12 | 25/2/17 | 25/5/14 |

to random sampling over the 44 datasets. Therefore, we consider the ratio $\frac{V_{active}}{V_{random}}$ where $V_{active}$ and $V_{random}$ are the ALC scores of active learning and random sampling, respectively. This gives us an indication of the relative improvement (or deterioration) the active learning schemes provide. We compute the ratios over all the datasets and visualize the outcomes with a box plot in Fig 5. The 25th, 50th and 75th percentiles are shown and the green crosses indict the average values of the ratios. We can observe that ENTROPY and EER may deliver satisfactory performances, while MAXER and CEER behave rather poorly. MLI achieves the highest ratio on one dataset, which means that MLI can improve most upon random sampling in some instances. Possibly more importantly, however, ENTROPY and EER may be considered safer: they may not reach the relative improvements that MLI does, but at least they also do not show dramatic decreases in performance. Even though random sampling strategy is expected to be less efficient than actual active learning algorithms, at times, it can perform so well in comparison to the latter. Similar observations have been made before in (Guyon et al., 2011) that random sampling is a runner-up in the active learning challenge.

Table 2 is divided into three different sections according to the ALC value achieved by random sampling. The first group, in which ALC scores range from 0.5 to 0.75, represents the datasets on which reaching good performance seems difficult. The second group, ranging from 0.75 to 0.90, corresponds to the datasets which have medium levels of difficulty for classification. The last group consists of the remaining datasets, which seem fairly easy to solve by a linear logistic classifier. We can see that random sampling surpasses all the other methods on the blood dataset. On the medium and easy datasets, random sampling does not achieve the best performances, which may indicate that we need only consider random sampling on relatively hard tasks. For the difficult classification datasets in the first group, ENTROPY, FIVR and MLI achieve comparable performances. FIVR performs best on the datasets in the first group while it performs poorly on the easy and medium datasets. In the second group, EVR obtains the best performance, while it underperforms in the last group. For the easy datasets, MMC, MLI and ENTROPY are slightly better than the other methods. ENTROPY also performs well on medium and hard datasets. The experiments demonstrate that uncertainty sampling is a robust active learning algorithm, regardless of the difficulty-level of the tasks.

Table 3 shows the average performance of all the methods over 80 subsets of the ImageNet database. A detailed description of the performance on each subset is shown in the Appendix. We can observe that, also in this setting, ENTROPY performs the best in terms of the four measures, i.e. average ALC, average ranking, win counts, and win/tie/loss. Interestingly, all other methods are outperformed by random sampling in terms of average ALC and average ranking. In conclusion, all but the simplest approach overall fail to outperform random sampling on this particular ImageNet database. This seems to indicate that more attention may have to be devoted to seeking safe, yet effective active learning algorithms (Loog & Yang, 2016).

### 3.4. Computational Cost Analysis

Computational cost can also be a critical issue when employing active learning methods. Table 4 assesses the average computational cost of selecting 40 unlabeled samples of each of the methods. All the experiments are constructed with MATLAB 9.1 on an Intel(R) Core 2.8GHz i7-4980HQ CPU PC with 16 GB memory. We test the computational cost on 8 datasets that vary in the numbers of instances and the feature dimensionalities. Clearly, random sampling, ENTROPY and MMC are the most efficient methods due to their simplicity. MLI also has a low computational burden compared with other algorithms. Error reduction and variance reduction have, on the other hand, a significantly higher computational cost than other methods. Both of them are especially significantly less efficient for handling high dimensionality datasets like 3vs5 and basehockey. The reason may be that both need to retrain the logistic regressor in every selection step over all the unlabeled instances and all possible labels, which is relatively time consuming especially in higher dimensions. We also note that EVR has the highest computational cost. This is because EVR has to repeatedly calculate the inverse of matrix, which is extremely computationally expensive. AAL has the second-highest computational cost since it also needs to compute the inverse matrix multiple times.

## 4. Discussion and Conclusion

This survey focuses on logistic regression because it is broadly applied and because of the fact that many active learning methods can be used in combination with this particular classifier. It should be clear, however, that some categories of active learning discussed in this work can also be used in combination with other types of classifiers. For instance, uncertainty sampling and error reduction approaches can be readily employed in combination with other probabilistic classifiers that can provide a posterior probability per sample, e.g. like naive Bayes (Roy & Mccallum, 2001). Especially in the two-class case, uncertainty sampling can already be applied as soon as one has a notation of distance to the decision boundary. A technique like minimum loss increase has also been studied in relation with SVMs (Hoi et al., 2008) and ridge regression (Huang et al., 2014). Maximum model change can also be used in combination with SVMs (Cai et al., 2017).

A Benchmark and Comparison of Active Learning for Logistic Regression

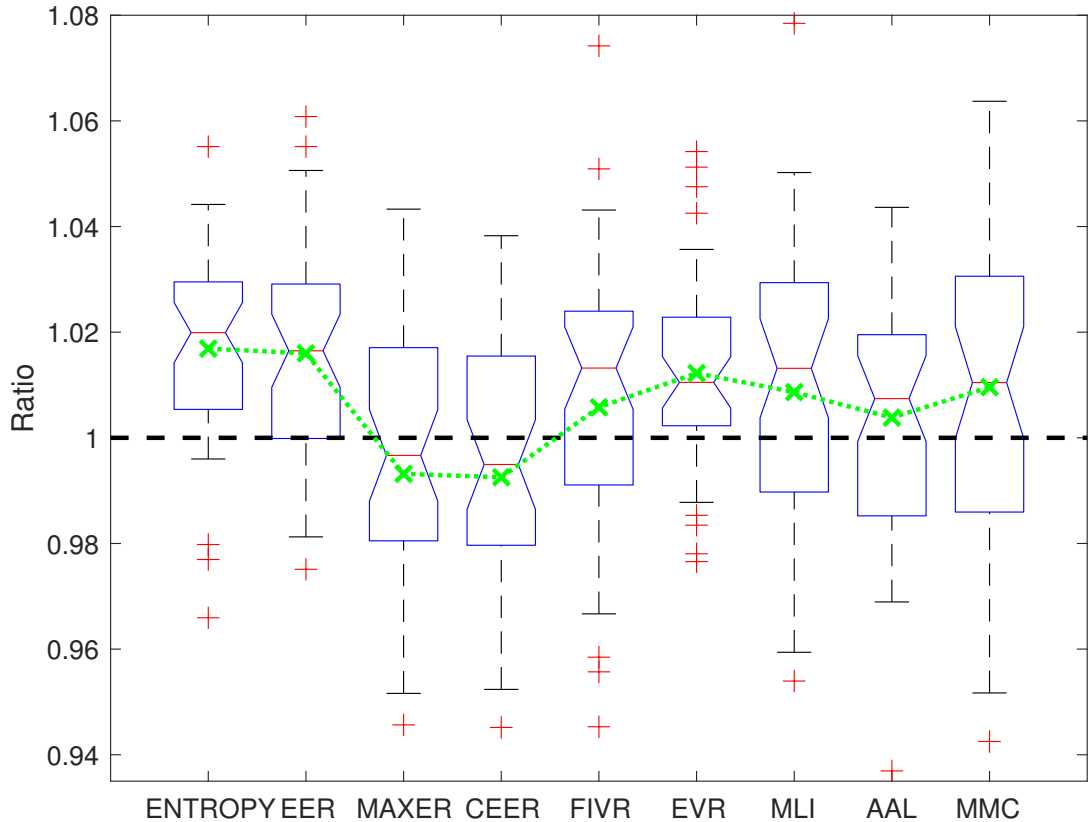

*Figure 5.* A box plot of the ratios of the ALC scores of active learning to that of random sampling over all the datasets. The green crosses represent the average values of the ratios. The black dashed line is at one, at the performance of random sampling.

On the other hand, there are also some active learning algorithms which are not easily combined with logistic regression. Examples are particular graph-based methods (Ji & Han, 2012; Ma et al., 2013) and methods that rely on model change with a closed-form estimate (Freytag et al., 2013) as these methods are specifically derived on the basis of Gaussian Processes. Other approaches rely on the notion of a version space or a margin (Tong & Koller, 2002; Kremer et al., 2014) and therefore can also not be combined readily with logistic regression.

More recently, quite some effort has gone into the study of scenarios that deviate to a smaller or larger extent from the standard myopic active learning setting that we focus on. The main research directions that we identified are multi-label active learning where every instance may have multiple labels simultaneously (Huang et al., 2015; Cherman et al., 2017), multi-task active learning in which various tasks are meant to be learned jointly (Harpale, 2012; Zhang, 2010), multiple instance active learning where human experts annotate an entire set that contains some samples instead of individual instances (Carbonneau et al., 2017; Zhang et al., 2010), cost-sensitive active learning where different samples have varying labeling costs (Persello et al., 2014; Wang et al., 2016), and active transfer learning which combines transfer learning and active learning (Gavves et al., 2015; Wang et al., 2014).

Finally, there are of course approaches in which deep learning and active learning come together. An original application is (Zhu & Bento, 2017) which proposed to use a generative adversarial network (GAN) to synthesize training instances for labeling instead of using real, observed samples. Another contribution, offering an original way of active labeling is (Huijser & van Gemert, 2017). In that work, a GAN is used to generate new images along a 1-dimensional query line and a human expert is asked, rather than to provide a label, to provide the point where the images change class.

In this paper, we compared current state-of-the-art active learning methods for logistic regression and pointed out their main similarities and dissimilarities. The experiments on the synthesis datasets and the large number of real-world datasets show some of the chief underlying characteristics of each of the active learning methods. On average, we would deem ENTROPY the most promising method. Though ENTROPY is a rather simplistic crite-

A Benchmark and Comparison of Active Learning for Logistic Regression

Table 3. Average performance of the Area under the Learning Curve on 80 subsets of the ImageNet database. The best performances are highlighted in bold face and surrounded with a box. "Mean" reports the average performance of the Area under the Learning Curve. "Average Ranking" shows the average ranking of compared methods. "Win counts" shows the total number of datasets on which one method achieves the best performances. "win/tie/loss" demonstrates the win/tie/loss counts of one method versus random sampling on all the datasets based on paired $t$-test at 95 percent significance level.

| Dataset | Random | ENTROPY | EER | MAXER | CEER | FIVR | EVR | MLI | AAL | MMC |
|---|---|---|---|---|---|---|---|---|---|---|
| Mean | 0.778 | **0.783** | 0.763 | 0.761 | 0.760 | 0.775 | 0.765 | 0.762 | 0.761 | 0.774 |
| Average Ranking | 3.96 | **2.58** | 6.54 | 6.66 | 6.78 | 4.40 | 6.53 | 7.22 | 6.03 | 4.31 |
| Win counts | 11 | **30** | 2 | 3 | 2 | 17 | 3 | 0 | 9 | 16 |
| win/tie/loss | - | **58/7/15** | 15/7/58 | 17/8/55 | 21/4/55 | 35/8/37 | 13/3/64 | 3/4/73 | 22/6/52 | 31/10/39 |

Table 4. Computational cost comparison of querying 40 unlabeled instances for each active learning method (in seconds)

| Dataset (#Ins #Fea) | Random | Entropy | EER | MAXER | CEER | FIVR | EVR | MLI | AAL | MMC |
|---|---|---|---|---|---|---|---|---|---|---|
| acute (120,6) | 0.006 | 0.015 | 0.520 | 0.502 | 0.824 | 0.085 | 0.719 | 0.530 | 0.171 | 0.017 |
| australian (690,14) | 0.013 | 0.030 | 6.794 | 6.619 | 8.618 | 0.830 | 11.287 | 4.794 | 11.848 | 0.044 |
| musk1 (476, 166) | 0.005 | 0.054 | 16.855 | 16.113 | 18.077 | 8.725 | 39.832 | 10.510 | 14.159 | 0.059 |
| hill (600, 100) | 0.006 | 0.045 | 16.842 | 16.534 | 19.355 | 5.011 | 34.041 | 8.121 | 16.025 | 0.051 |
| mushrooms (1000, 112) | 0.007 | 0.029 | 15.179 | 15.313 | 17.706 | 7.787 | 89.594 | 9.686 | 46.020 | 0.110 |
| letterEF (1543, 16) | 0.006 | 0.031 | 25.400 | 25.382 | 28.850 | 1.129 | 44.214 | 11.635 | 203.500 | 0.082 |
| 3vs5 (1500, 784) | 0.006 | 0.186 | 219.578 | 219.838 | 230.834 | 609.544 | 1806.288 | 55.369 | 216.266 | 0.247 |
| basehockey (1993, 500) | 0.007 | 1.132 | 1133.409 | 1122.366 | 1139.517 | 289.424 | 2060.489 | 165.871 | 2251.607 | 0.719 |
| Mean | 0.007 | 0.190 | 179.322 | 177.834 | 182.973 | 115.317 | 510.808 | 33.315 | 344.949 | 0.166 |

rion and quite short-sighted when picking instances, it does outperform the min-max view approach, variance reduction methods and maximum model change algorithm in our experiments. Uncertainty sampling was first proposed in 1994, which may indicate that, in some sense, little progress has been made since then. MLI demonstrates its advantage in querying the representative instances on the synthesis data Synth2. A possible downside of expected error reduction approaches is the high computational cost it incurs. Variance reduction approaches and MLI suffer the same problem. How to speed up these methods is definitely a worthwhile problem for further research.

Overall, on the positive side, we can conclude that active learning can indeed provide improved performance over random sampling, most certainly if we consider the whole ensemble of active learners in this work. This, however, also seems to be a negative aspect. On its own, none of these methods can prevent becoming worse than random sampling. While this seems impossible anyway for every single instantiation of a problem, our results indicate that it does not even hold in the average. That is, for every active learner there are (real-world) datasets on which the active learner performs significantly worse than random sampling, even when averaged over multiple runs. Finding active learning methods that are, in some sense, safe and yet give significant performance gains at times, still seems to be the challenge ahead (cf. (Attenberg & Provost, 2011; Loog et al., 2016; Loog & Yang, 2016; Beygelzimer et al., 2010; Settles, 2011)).

# Acknowledgements
We thank the associate editor and the three reviewers for their helpful comments that improved the manuscript significantly. We also thank Stephanie Tan (Delft University of Technology) for her kind help and suggestions concerning spelling and grammar.

# Appendix

# A Benchmark and Comparison of Active Learning for Logistic Regression

Table A.1: Performance comparison on the Area under the Learning Curve (Accuracy). The higher the score, the better the performance. For each data set, the best performances and its comparable competitors are highlighted in bold face and surrounded with a box. Average performance of all the active learning methods are also reported as "Mean". "Average Ranking" shows the average ranking of compared methods. "Win counts" shows the total number of datasets on which one method achieves the best performances. "win/tie/loss" demonstrates the win/tie/loss counts of one method versus random sampling on all the datasets based on paired $t$-test at 95 percent significance level.

| Dataset | Random | ENTROPY | EER | MAXER | CEER | FIVR | EVR | MLI | AAL | MMC |
|---|---|---|---|---|---|---|---|---|---|---|
| Egyptian | 0.554 | 0.550 | **0.559** | 0.546 | 0.545 | 0.546 | 0.555 | 0.544 | 0.553 | 0.533 |
| Tabby | 0.565 | 0.569 | 0.569 | 0.568 | 0.570 | 0.564 | **0.586** | 0.559 | 0.563 | 0.570 |
| Siamese | 0.616 | 0.622 | 0.621 | 0.615 | 0.618 | **0.630** | **0.632** | 0.611 | 0.628 | 0.614 |
| Persian | 0.621 | 0.629 | 0.640 | 0.628 | 0.633 | **0.661** | 0.650 | 0.606 | 0.638 | 0.619 |
| umbrella vs. ball | 0.622 | 0.614 | 0.604 | 0.572 | 0.577 | **0.626** | 0.592 | 0.604 | 0.554 | 0.599 |
| computermouse vs. helmet | **0.638** | 0.624 | 0.617 | 0.599 | 0.601 | 0.620 | 0.617 | 0.624 | 0.592 | 0.630 |
| scissors vs. cellphone | 0.638 | **0.641** | 0.624 | 0.616 | 0.623 | 0.633 | 0.615 | 0.619 | 0.636 | 0.632 |
| bottle vs. cellphone | 0.638 | **0.645** | 0.616 | 0.617 | 0.616 | 0.624 | 0.631 | 0.629 | 0.624 | 0.631 |
| ewer vs. knife | **0.644** | 0.629 | 0.610 | 0.586 | 0.586 | 0.588 | 0.631 | 0.637 | 0.592 | 0.630 |
| spoon vs. telephone | **0.649** | 0.641 | 0.632 | 0.623 | 0.623 | 0.635 | 0.639 | 0.642 | 0.618 | 0.626 |
| catrabbit | 0.652 | 0.663 | 0.667 | 0.652 | 0.659 | 0.648 | 0.659 | 0.657 | **0.669** | 0.637 |
| Tiger | 0.655 | 0.660 | 0.667 | 0.658 | 0.670 | 0.645 | **0.672** | 0.629 | 0.658 | 0.637 |
| bottle vs. spoon | 0.666 | 0.680 | 0.648 | 0.667 | 0.657 | 0.635 | 0.649 | 0.639 | **0.687** | 0.635 |
| calculator vs. cellphone | 0.672 | **0.674** | 0.649 | 0.652 | 0.634 | 0.650 | 0.656 | 0.659 | 0.639 | 0.644 |
| teapot vs. lightbulb | **0.678** | 0.676 | 0.667 | 0.620 | 0.646 | 0.668 | 0.658 | 0.659 | 0.640 | **0.678** |
| spoon vs. cartire | **0.679** | 0.675 | 0.658 | 0.647 | 0.659 | 0.648 | 0.655 | 0.672 | 0.650 | 0.671 |
| flag vs. tower | **0.682** | 0.679 | 0.667 | 0.664 | 0.670 | 0.678 | 0.673 | 0.661 | 0.679 | 0.667 |
| rifle vs. eyeglasses | 0.701 | **0.713** | 0.692 | 0.684 | 0.683 | 0.686 | 0.667 | 0.672 | 0.693 | 0.684 |
| truck vs. boat | 0.714 | 0.728 | 0.695 | 0.681 | 0.687 | **0.736** | 0.704 | 0.688 | 0.662 | 0.714 |
| table vs. cellphone | **0.718** | 0.713 | 0.689 | 0.673 | 0.670 | 0.674 | 0.682 | 0.698 | 0.651 | 0.680 |
| motorcycle vs. baseballbat | **0.719** | 0.717 | 0.681 | 0.658 | 0.658 | 0.664 | 0.700 | 0.674 | 0.666 | 0.678 |
| skunk vs. umbrella | 0.721 | **0.724** | 0.699 | 0.699 | 0.683 | 0.720 | 0.698 | 0.695 | 0.702 | 0.701 |
| catpanda | 0.725 | **0.742** | 0.726 | 0.731 | 0.728 | 0.691 | 0.736 | 0.703 | **0.743** | 0.681 |
| apple vs. cup | 0.725 | 0.728 | 0.723 | 0.715 | 0.707 | **0.735** | 0.708 | 0.714 | 0.718 | 0.722 |
| sheep vs. skunk | 0.726 | 0.725 | 0.719 | 0.718 | 0.716 | 0.724 | 0.716 | 0.708 | **0.735** | 0.705 |
| motorcycle vs. bridge | 0.732 | **0.737** | 0.661 | 0.637 | 0.634 | 0.727 | 0.699 | 0.697 | 0.603 | 0.734 |
| bike vs. spoon | 0.754 | **0.769** | 0.733 | 0.751 | 0.750 | 0.744 | 0.750 | 0.736 | 0.765 | 0.753 |
| bathtub vs. basketball_hoop | **0.756** | **0.756** | 0.736 | 0.734 | 0.733 | 0.749 | 0.734 | 0.734 | **0.756** | 0.741 |
| washingmachine vs. cup | 0.759 | **0.772** | 0.736 | 0.729 | 0.727 | 0.754 | 0.736 | 0.746 | 0.754 | 0.748 |
| piano vs. scissors | **0.763** | 0.755 | 0.721 | 0.728 | 0.728 | 0.713 | 0.725 | 0.743 | 0.731 | 0.756 |
| lama vs. kangaroo | 0.766 | 0.771 | 0.779 | 0.780 | 0.788 | 0.790 | 0.782 | 0.764 | 0.766 | **0.793** |
| catelepant | 0.768 | 0.795 | 0.781 | 0.788 | 0.791 | 0.765 | 0.789 | 0.762 | **0.797** | 0.763 |
| horse vs. windmill | 0.770 | 0.771 | 0.728 | 0.741 | 0.722 | **0.795** | 0.754 | 0.750 | 0.737 | 0.769 |
| washingmachine vs. umbrella | 0.771 | **0.785** | 0.770 | 0.767 | 0.767 | 0.771 | 0.762 | 0.762 | 0.776 | 0.780 |
| bear vs. flower | 0.772 | **0.775** | 0.737 | 0.708 | 0.700 | 0.770 | 0.748 | 0.765 | 0.734 | 0.726 |
| piano vs. ball | 0.772 | **0.781** | 0.755 | 0.773 | 0.760 | 0.779 | 0.753 | 0.754 | 0.759 | 0.779 |
| scorpion vs. scissors | 0.773 | 0.772 | 0.763 | 0.759 | 0.759 | 0.739 | 0.759 | 0.750 | **0.776** | 0.749 |
| tomato vs. umbrella | 0.774 | **0.785** | 0.765 | 0.765 | 0.764 | 0.778 | 0.762 | 0.752 | 0.769 | 0.751 |
| buildings vs. lamp | 0.776 | **0.783** | 0.757 | 0.761 | 0.759 | 0.775 | 0.757 | 0.753 | 0.769 | 0.762 |
| billiards vs. umbrella | 0.779 | **0.788** | 0.742 | 0.728 | 0.726 | 0.759 | 0.746 | 0.762 | 0.713 | 0.770 |
| binder vs. grand_piano | 0.790 | 0.785 | 0.774 | 0.779 | 0.782 | 0.805 | 0.775 | 0.772 | 0.782 | **0.819** |
| chandelier vs. baseballglove | 0.791 | 0.797 | 0.768 | 0.755 | 0.762 | 0.798 | 0.791 | 0.779 | 0.740 | **0.806** |
| knob vs. giraffe | 0.791 | **0.805** | 0.765 | 0.755 | 0.758 | 0.789 | 0.774 | 0.775 | 0.770 | 0.798 |
| palmtree vs. rifle | 0.799 | 0.808 | 0.776 | 0.792 | 0.786 | 0.798 | 0.767 | 0.777 | **0.811** | 0.768 |
| bird vs. tombstone | 0.805 | 0.804 | 0.807 | 0.798 | 0.784 | 0.804 | 0.778 | 0.807 | 0.758 | **0.813** |
| motorcycle vs. buildings | 0.808 | **0.820** | 0.777 | 0.758 | 0.762 | 0.791 | 0.794 | 0.780 | 0.787 | 0.815 |
| flower vs. ewer | 0.815 | 0.812 | 0.805 | 0.795 | 0.793 | **0.834** | 0.806 | 0.804 | 0.797 | 0.816 |
| dog vs. headphone | 0.817 | **0.825** | 0.795 | 0.793 | 0.786 | 0.816 | 0.810 | 0.813 | 0.803 | 0.823 |
| horse vs. shoe | 0.819 | 0.823 | 0.805 | 0.791 | 0.795 | **0.830** | 0.804 | 0.817 | 0.815 | 0.826 |
| umbrella vs. gorilla | 0.819 | **0.832** | 0.812 | 0.828 | 0.827 | 0.812 | 0.809 | 0.807 | **0.833** | 0.816 |

Table A.1 – *Continued from previous page*

| Dataset | Random | ENTROPY | EER | MAXER | CEER | FIVR | EVR | MLI | AAL | MMC |
|---|---|---|---|---|---|---|---|---|---|---|
| chandelier vs. mushroom | 0.820 | **0.829** | 0.808 | 0.810 | 0.800 | 0.803 | 0.803 | 0.794 | 0.810 | 0.810 |
| bottle vs. boat | 0.829 | **0.836** | 0.812 | 0.820 | 0.818 | **0.836** | 0.813 | 0.819 | 0.828 | 0.821 |
| lighthouse vs. bike | 0.830 | **0.845** | 0.817 | 0.808 | 0.799 | 0.842 | 0.811 | 0.809 | 0.805 | 0.827 |
| bridge vs. skunk | 0.831 | **0.842** | 0.784 | 0.790 | 0.780 | 0.840 | 0.792 | 0.806 | 0.799 | 0.824 |
| lama vs. can_soda | 0.833 | **0.841** | 0.813 | 0.791 | 0.812 | 0.840 | 0.811 | 0.820 | 0.751 | 0.836 |
| lightbulb vs. giraffe | 0.843 | 0.850 | 0.831 | 0.824 | 0.826 | **0.865** | 0.841 | 0.827 | 0.826 | 0.863 |
| cerealbox vs. skunk | 0.846 | 0.859 | 0.845 | 0.858 | 0.855 | **0.865** | 0.844 | 0.830 | 0.853 | 0.854 |
| dog vs. stapler | 0.847 | 0.848 | 0.827 | 0.833 | 0.828 | 0.841 | 0.831 | 0.841 | 0.833 | **0.850** |
| traffic_light vs. chimp | 0.848 | 0.856 | 0.819 | 0.835 | 0.827 | **0.858** | 0.822 | 0.835 | 0.801 | 0.850 |
| bridge vs. washingmachine | 0.849 | **0.864** | 0.839 | 0.844 | 0.850 | 0.857 | 0.838 | 0.834 | 0.837 | 0.839 |
| octopus vs. scissors | **0.851** | **0.850** | 0.820 | 0.821 | 0.821 | 0.800 | 0.828 | 0.832 | 0.824 | 0.843 |
| bear vs. fireextinguisher | 0.853 | 0.861 | 0.837 | 0.823 | 0.820 | 0.856 | 0.842 | 0.845 | 0.834 | **0.864** |
| cannon vs. tombstone | 0.857 | 0.868 | 0.857 | 0.864 | 0.863 | 0.852 | 0.852 | 0.854 | 0.863 | **0.871** |
| aeroplane vs. sheep | 0.857 | 0.871 | 0.861 | 0.857 | 0.866 | 0.878 | 0.861 | 0.850 | 0.862 | **0.881** |
| frying_pan vs. bear | 0.864 | 0.865 | 0.828 | 0.816 | 0.812 | **0.880** | 0.831 | 0.845 | 0.801 | 0.875 |
| chandelier vs. kangaroo | 0.865 | 0.870 | 0.831 | 0.814 | 0.808 | **0.878** | 0.827 | 0.841 | 0.787 | 0.872 |
| spoon vs. tombstone | 0.867 | 0.874 | 0.857 | 0.870 | **0.877** | 0.845 | 0.852 | 0.850 | 0.853 | 0.872 |
| cerealbox vs. monitor | 0.871 | 0.876 | 0.865 | 0.866 | 0.866 | **0.882** | 0.848 | 0.845 | 0.848 | 0.881 |
| butterfly vs. cerealbox | 0.873 | **0.883** | 0.873 | 0.873 | 0.878 | 0.870 | 0.868 | 0.845 | 0.873 | 0.879 |
| headphone vs. people | 0.875 | **0.891** | 0.862 | 0.868 | 0.877 | 0.874 | 0.877 | 0.860 | 0.880 | **0.891** |
| tenniscourt vs. ladder | 0.879 | 0.886 | 0.886 | **0.892** | 0.891 | 0.890 | 0.885 | 0.878 | 0.889 | 0.877 |
| people vs. computermouse | 0.895 | 0.906 | 0.890 | 0.908 | 0.909 | 0.911 | 0.889 | 0.894 | 0.904 | **0.914** |
| firetruck vs. bathtub | 0.899 | 0.903 | 0.879 | 0.904 | 0.902 | 0.910 | 0.897 | 0.892 | 0.897 | **0.915** |
| keyboard vs. bonsai | 0.900 | 0.913 | 0.906 | **0.921** | 0.918 | 0.906 | 0.886 | 0.879 | 0.914 | 0.913 |
| keyboard vs. bonsai | 0.900 | 0.913 | 0.906 | **0.921** | 0.918 | 0.906 | 0.886 | 0.879 | 0.914 | 0.913 |
| skyscraper vs. bonsai | 0.901 | 0.903 | 0.867 | 0.870 | 0.869 | 0.907 | 0.884 | 0.882 | 0.896 | **0.909** |
| teapot vs. tree | 0.911 | **0.918** | **0.920** | 0.914 | 0.911 | **0.920** | 0.895 | 0.895 | 0.894 | 0.914 |
| frying_pan vs. microwave | 0.919 | 0.907 | 0.868 | 0.906 | 0.898 | 0.911 | 0.881 | 0.895 | 0.892 | **0.921** |
| flashlight vs. tombstone | 0.922 | 0.922 | 0.922 | 0.925 | 0.926 | 0.930 | 0.919 | 0.909 | 0.916 | **0.939** |
| tree vs. dinosaur | 0.936 | 0.941 | 0.939 | 0.944 | **0.946** | 0.943 | 0.938 | 0.938 | 0.942 | **0.946** |
| Mean | 0.778 | **0.783** | 0.763 | 0.761 | 0.760 | 0.775 | 0.765 | 0.762 | 0.761 | 0.774 |
| Average Ranking | 3.96 | **2.58** | 6.54 | 6.66 | 6.78 | 4.40 | 6.53 | 7.22 | 6.03 | 4.31 |
| Win counts | 11 | **30** | 2 | 3 | 2 | 17 | 3 | 0 | 9 | 16 |
| win/tie/loss | - | **58/7/15** | 15/7/58 | 17/8/55 | 21/4/55 | 35/8/37 | 13/3/64 | 3/4/73 | 22/6/52 | 31/10/39 |